\documentclass{article}

\usepackage[dblblindworkshop, preprint]{neurips_2026}

\usepackage[utf8]{inputenc}
\usepackage[T1]{fontenc}
\usepackage{hyperref}
\usepackage{url}
\usepackage{booktabs}
\usepackage{amsmath}
\usepackage{amsfonts}
\usepackage{nicefrac}
\usepackage{microtype}
\usepackage{xcolor}
\usepackage{graphicx}

\title{Best-of-$N$ Test-Time Steering Can Lead to Reward Hacking in
Vision-Language-Action Models}

\workshoptitle{Workshop on Physical World AI}

\author{
Vanshaj Khattar$^{1}$,
Jing Liu$^{2}$,
Toshiaki Koike-Akino$^{2}$,
Ming Jin$^{1}$,
Ye Wang$^{2}$
\\[0.5em]
{\small $^{1}$Virginia Tech, $^{2}$Mitsubishi Electric Research Laboratories}
}

\begin{document}

\maketitle

\begin{abstract}
Vision-language-action (VLA) models have shown promise for general-purpose robotic manipulation, but adapting to changing environmental conditions still remains challenging.
Recent test-time steering methods have shown promise in enabling adaptive VLA policies by utilizing additional computation during deployment. In this paper, we consider best-of-$N$ test-time steering for VLA models, where multiple action candidates are sampled from a fixed policy and a verifier is used to select the best one during deployment. 
We show that increasing $N$ does not always improve performance when the verifier is inaccurate, and can favor candidates whose scores exploit prediction errors, a form of reward hacking. 
We study this effect on 60 LIBERO-Safety task-level cases using a $\pi_{0.5}$ policy, with ten repetitions per case.
We introduce controlled errors in a MuJoCo trajectory verifier and show that as $N$ increases, selected candidate rewards diverge from the accurate verifier rewards, leading to reward hacking.
We also show that verifier calibration and softmax-based candidate selection can mitigate reward hacking in some settings.
\end{abstract}

\section{Introduction}

Vision-language-action (VLA) models map visual observations and language instructions to robot actions and have shown broad manipulation capabilities \citep{brohan2023rt2,kim2024openvla,pi2025pi05}.
However, their performance can deteriorate when deployment conditions differ from those seen during training \citep{fei2025liberoplus}.
There are two ways to improve the generalization capabilities of VLAs.
The first is to collect additional demonstrations and retrain or fine-tune the policy; however, this can be computationally expensive \citep{lu2025vla, kim2025fine}.
Another widely used and less computationally expensive approach is test-time steering, where additional computation during execution is used to steer the VLA actions \citep{nakamoto2024vgps,dai2025rover} or select the best one from a few sampled candidates \citep{kwok2025robomonkey,zhang2026veritas}.

Recent test-time steering methods such as RoboMonkey and VERITAS have shown promising results with best-of-$N$ (BoN) test-time steering, which samples multiple action candidates and uses a verifier to select the best one for execution \citep{kwok2025robomonkey,zhang2026veritas}.
However, its benefit depends on the verifier's ability to rank those candidates.
In practice, perception errors, inaccurate predictions of physical motion, or an imperfect scoring rule can cause a poor action to receive a high score.

This problem has been identified in large language models (LLMs), where a reward model or verifier is used to pick the best of several sampled responses.
\cite{gao2023overoptimization} showed that when best-of-$N$ selection uses an imperfect reward model, the true quality of the selected response first improves and then drops as $N$ grows.
Similar effects have been shown when selecting solutions to math problems with a learned verifier~\cite{cobbe2021training} and when selecting code with unit tests~\cite{stroebl2026limits}, where wrong answers that pass the check limit the benefit of more samples.
Recent work explains this pattern in theory and shows that it is expected for best-of-$N$ selection when $N$ is large~\cite{huang2025bestofn, khalaf2025inference}, and terms it inference-time reward hacking.
These works also suggest ways to mitigate this effect, such as down-weighting candidates with uncertain scores~\cite{huang2025bestofn} or using softmax-based selection instead of always taking the top-scoring candidate~\citep{khalaf2025inference,wang2026slop}.

Similar effects have recently been reported in robotics literature.
\citet{luo2026critichack} showed that optimizing robot policies against a learned visual critic can increase failures in which the robot acts on the wrong object, because the critic scores some of these failures above real successes.
\citet{bakshi2026goodhart} showed that sampling-based planning with rewards learned from human preferences can reduce the true task return by up to 76\%, and that best-of-$N$ selection with such a reward can perform worse than taking a single action.
A recent survey of verifiers for robot policies notes that how far a policy can be optimized against a verifier before its score stops tracking the task is still largely unexplored~\citep{wan2026nofreechecker}.
These studies use learned scores and focus on task success. In contrast, we use a MuJoCo simulator whose errors we control, and a proxy reward function. We measure both safety violations and task completion.

This raises the central question of our study: when the verifier is imperfect, does additional search still improve robot behavior, or can it reduce task performance and safety?
We study this question on the LIBERO-Safety benchmark~\cite{cui2026liberosafety}, which requires robots to complete manipulation tasks while following constraints on how objects are grasped and avoiding contact with nearby obstacles and simulated human hands.
In these tasks, task progress and safety can conflict: an action can move an object toward its goal while violating a safety constraint.
We keep the VLA policy and the execution environment fixed, and add controlled errors only inside the verifier, either by adding noise to the predicted object positions or by changing the simulated dynamics.
For each selected action, we compare the \emph{proxy score} from the perturbed verifier with its \emph{clean score} from the unperturbed simulator.
\textbf{We find that best-of-$N$ steering of VLA policies is prone to inference-time reward hacking: }as $N$ increases, the gap between predicted and clean reward grows, and for some tasks the gains in safe completion seen at small $N$ are lost while violations increase.

\begin{figure}[!t]
  \centering
  \includegraphics[width=\linewidth]{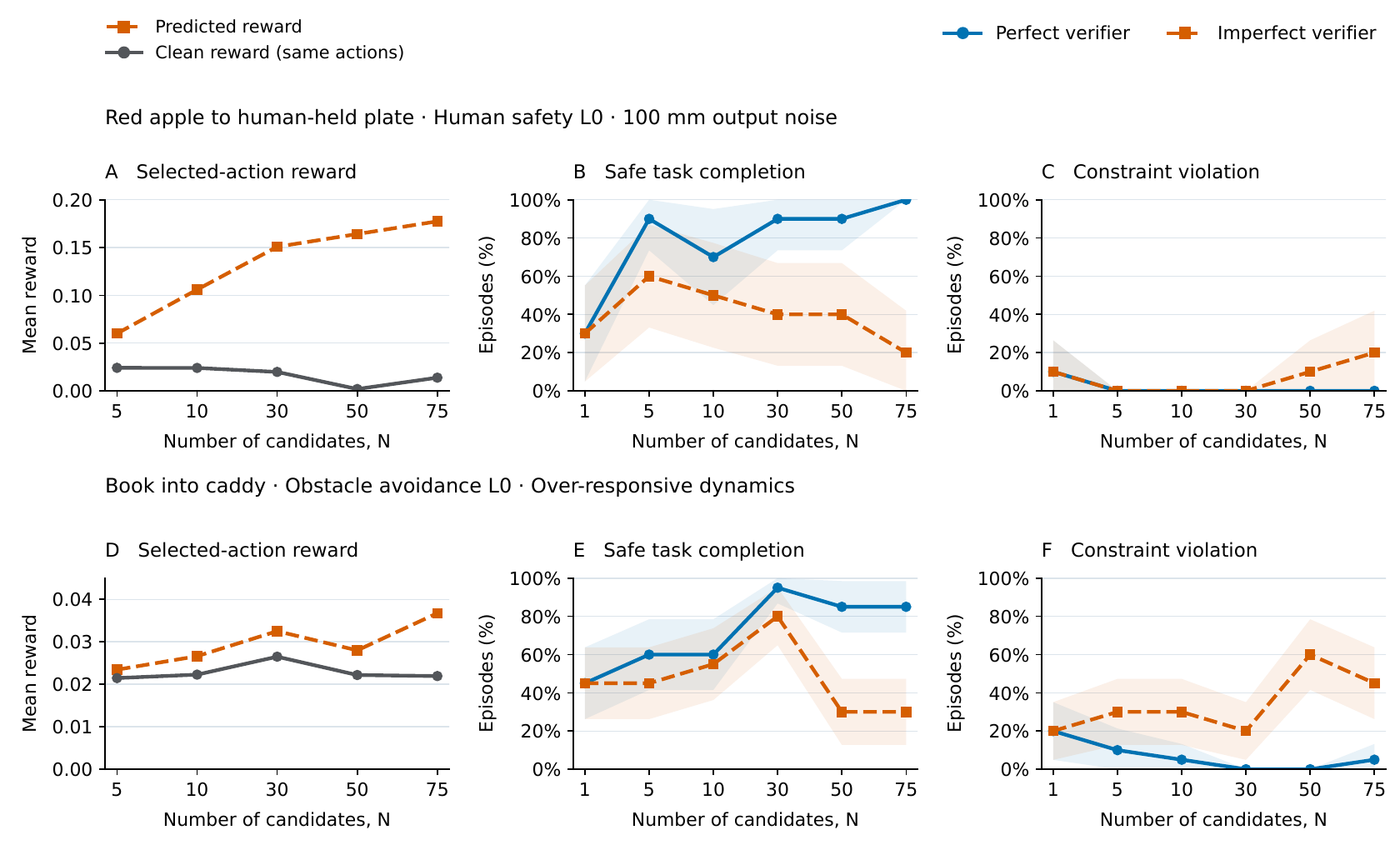}
  \caption{More candidates raise predicted reward but not true quality when the verifier is imperfect. Top: red apple placement with 100 mm noise in predicted object positions.
Bottom: book placement with over-responsive simulated dynamics. (A,~D)~Predicted and clean reward of the same selected actions. (B,~E)~Safe task completion and (C,~F)~constraint violations, compared with a perfect verifier. Bands show pointwise 90\% Wilson intervals over 10 episodes per point.}
  \label{fig:opening}
\end{figure}

\section{\texorpdfstring{Best-of-$N$}{Best-of-N} steering with an imperfect verifier}

At replanning step $t$, the VLA samples $N$ candidate action chunks $a_t^{1:N}=\{a_t^1,\ldots,a_t^N\}$ from $\pi(\cdot\mid o_t,l)$, given the current observation $o_t$ and language instruction $l$.
A verifier scores each candidate using a predictor $\hat f$ to estimate the trajectory that $a_t^i$ would produce over a short horizon, and from this trajectory computes a safety score $\hat S_i\in[-1,0]$ and a task-progress score $\hat T_i\in[-1,1]$ (Section~\ref{sec:scores}).
The scores are combined, and the best candidate is selected using:
\begin{equation}
  \hat R_i := w_S\hat S_i+w_T\hat T_i,
  \qquad
  i_N^* := \arg\max_{i\in\{1,\ldots,N\}}\hat R_i ,
  \label{eq:bon}
\end{equation}
where $w_S$ and $w_T$ set the relative weight of safety and task progress.
We use $w_S=w_T=0.5$ in the main experiments.
The selected chunk $a_t^{i_N^*}$ is executed for five control steps before replanning.

To measure how good the selected action actually is, we score it again with the unperturbed MuJoCo simulator.
We call this score the \emph{clean reward} $R_i^{\mathrm{clean}}$.
It uses the same short-horizon scoring rule as $\hat R_i$.
A \emph{perfect verifier} is one that selects candidates using $R_i^{\mathrm{clean}}$ directly.
We define
\begin{equation}
  P_N := \mathbb{E}\!\left[\hat R_{i_N^*}\right],\qquad
  G_N := \mathbb{E}\!\left[R_{i_N^*}^{\mathrm{clean}}\right],
  \label{eq:hacking}
\end{equation}
where $P_N$ is the expected proxy reward assigned by the verifier to the selected candidate, and $G_N$ is the expected clean reward assigned to that same candidate by the unperturbed simulator.
The expectation is taken over evaluation tasks, replanning steps, and candidate samples.

We say that reward hacking is shown when the gap between proxy and clean reward grows, and the increase in $P_N$ is not matched by clean reward or closed-loop outcomes.
The clearest case is when, for two candidate counts $N_1<N_2$, proxy reward increases while clean reward does not: $P_{N_2}>P_{N_1}$ and $G_{N_2}\leq G_{N_1}$. 
For example, in Figure \ref{fig:opening}A, the proxy reward keeps rising as $N$ increases from 5 to 50, while the clean reward of the same actions falls.

\subsection{Safety and task scores}
\label{sec:scores}

The verifier scores each candidate from its predicted rollout of $h=5$ control steps.
Let $\hat v_i$ be the number of constraint violations in the predicted rollout, counted over all steps and all constraints.
The safety score is
\begin{equation}
  \hat S_i=-\operatorname{clip}(\hat v_i/h,\,0,\,1).
  \label{eq:safety}
\end{equation}
A candidate with no predicted violations scores $0$, and each violation gives a score of $-1$.

\paragraph{Task score.}
The proxy task score $\hat T_i$ is computed from the same predicted rollout.
Each task is specified by a list of goal conditions, such as putting the apple on the plate, and the score measures progress toward the first condition that is not yet satisfied.
Its dense scoring part $\hat D_i$ rewards moving the gripper toward the target object and grasping it, and, once the object is held, lifting it and moving it toward its goal location.
We add the simulator's sparse reward summed over the rollout, $\hat r_i^{\mathrm{sum}}$, and a success bonus $\hat s_i\in\{0,1\}$ that equals $1$ if the predicted rollout completes the task:
\begin{equation}
  \hat T_i=\operatorname{clip}\!\left(
  \frac{\hat D_i+\hat r_i^{\mathrm{sum}}+\hat s_i}{5},\,-1,\,1\right),
  \label{eq:task}
\end{equation}
where dividing by $5$ scales the score to a range similar to that of $\hat S_i$.
Because $\hat r_i^{\mathrm{sum}}$ and $\hat s_i$ both come from the verifier's simulator, they are also predictions and are affected by the verifier's errors.
The clean reward $R_i^{\mathrm{clean}}$ applies the same rules in the unperturbed simulator, with $w_S=w_T=0.5$ in all experiments.
Appendix~\ref{app:task-score} lists every term in detail.

\section{Empirical Studies}

\paragraph{Setup.}
We evaluate $\pi_{0.5}$ \citep{pi2025pi05} using the LIBERO-Safety-finetuned checkpoint released with the benchmark \citep{cui2026liberosafety}. \footnote{\url{https://huggingface.co/LIBERO-Safety/pi05_libero_safety}}
We evaluate 60 task-level cases from four task suites: affordance-aware grasping, human-robot interaction, tabletop obstacle avoidance, and hand-object avoidance.
Each suite contains five task instructions evaluated at three levels of difficulty (L0-L2), giving 15 tasks per suite.
Episodes run for at most 200 control steps.
At each replanning step, the policy samples $N\in\{1,5,10,30,50,75\}$ action chunks from its current observation. 
Episode comparisons use ten policy-sampling repetitions per case, condition, and $N$.

\paragraph{Imperfect verifier details.}
We compare an unperturbed predictor against three imperfect predictors.
\emph{Output noise} adds independent Gaussian noise with standard deviation $0.1$~m to predicted object $X$, $Y$, and $Z$ coordinates.
\emph{Under-responsive} and \emph{over-responsive} dynamics make the robotic arm respond more weakly or strongly to the same commands.
These settings modify actuator strength, damping, friction, armature, gripper strength, and object mass in the MuJoCo simulator.
The simulator state is restored before execution, so the errors affect only candidate scoring.

\paragraph{Episode metrics.}
Task completion means that the benchmark's full task goal was reached, regardless of safety.
A violation means that at least one benchmark constraint was violated.
\emph{Safe task completion} requires both task completion and no violation in the evaluated episode.
We report pointwise 90\% Wilson intervals.
Mean predicted reward curves show means over replanning decisions without confidence intervals.

\subsection{RQ1: Does verifier error produce reward hacking behavior?}
\label{sec:rq1}

\begin{figure}[t]
  \centering
  \includegraphics[width=\linewidth]{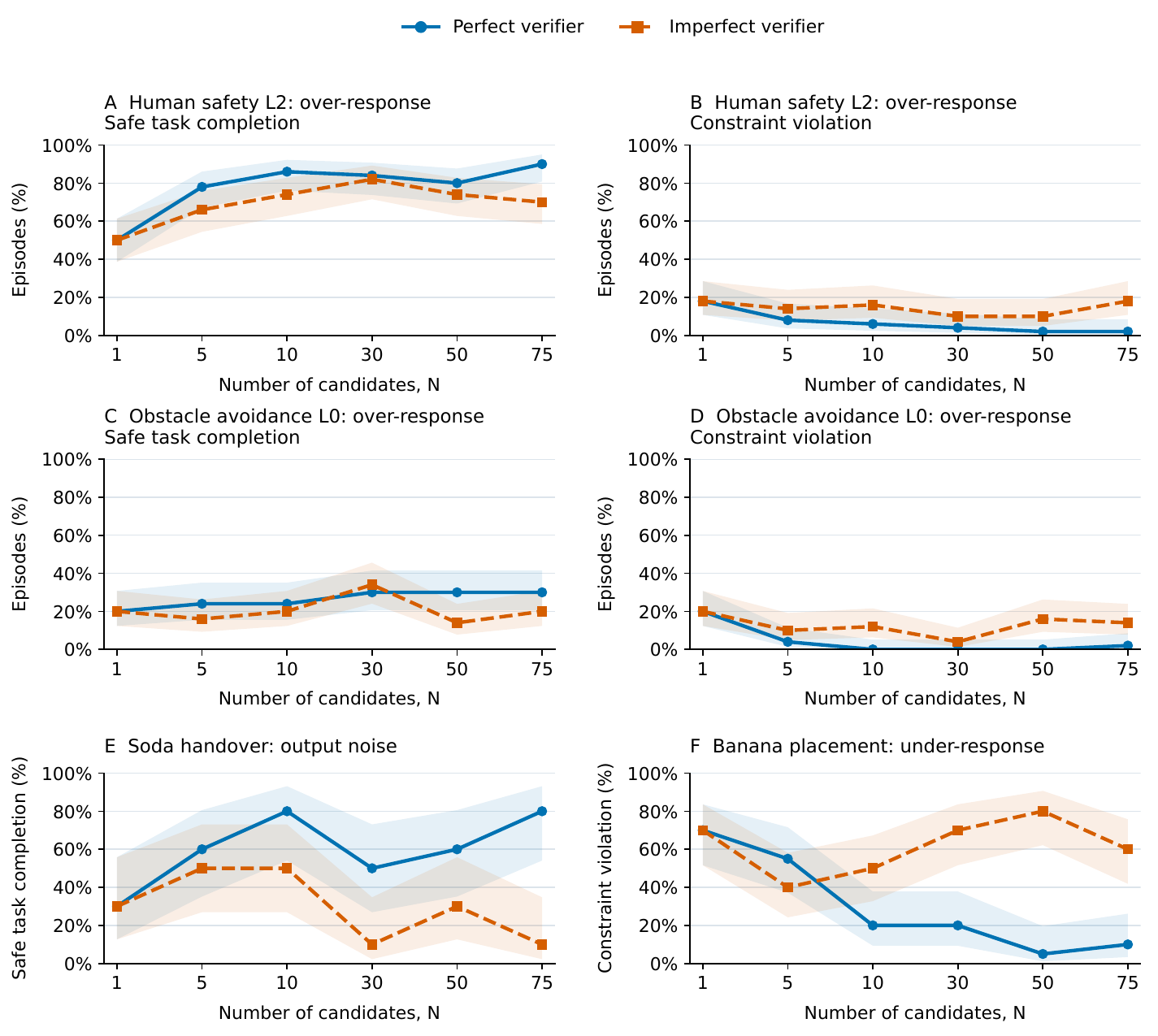}
  \caption{(A--D) Over-responsive dynamics: safe task completion (A, C) and constraint violations (B, D) for human safety L2 tasks (A, B) and obstacle avoidance L0 (C, D). (E) Safe task completion for soda handover with noise in predicted object positions. (F) Constraint violations for banana placement with under-responsive simulated dynamics. Bands show 90\% Wilson confidence intervals.}
  \label{fig:rq1-suites}
  \label{fig:rq1-tasks}
\end{figure}

Figure \ref{fig:opening} shows that imperfect verification can favor actions with higher predicted rewards but lower clean rewards and worse task completion or higher constraint violations.
Figure~\ref{fig:rq1-suites} shows how the executed outcomes change in additional task groups and individual tasks.

\begin{figure}[!htbp]
  \centering
  \includegraphics[width=\linewidth]{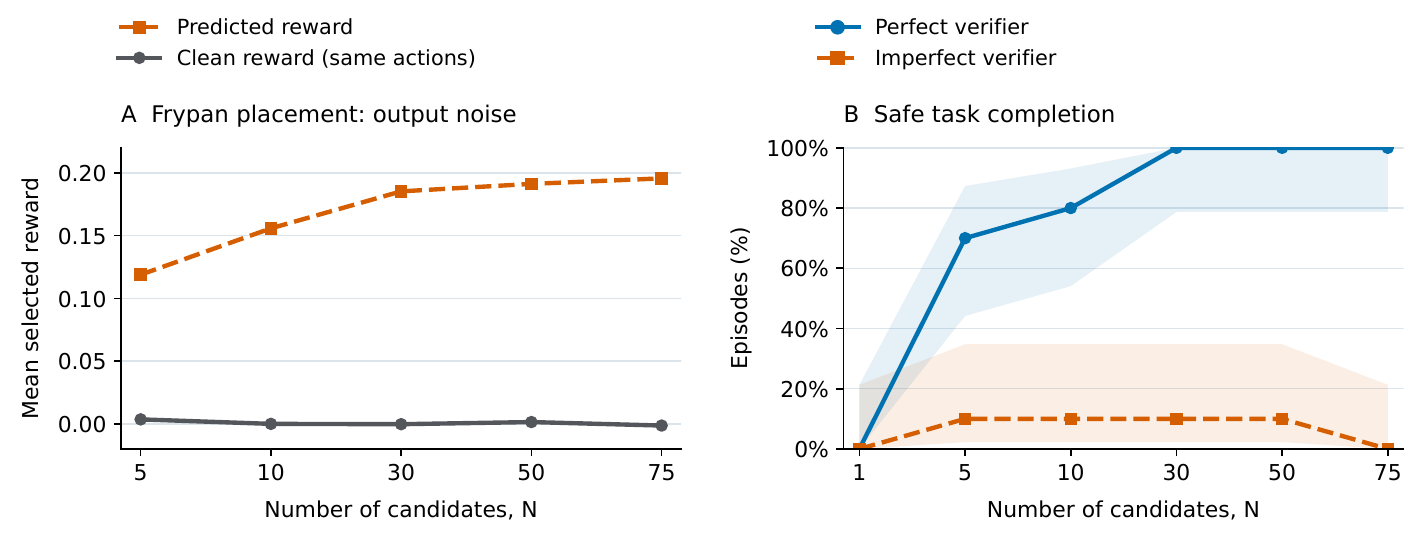}
  \caption{Frypan placement task with noise in predicted object positions. (A) Predicted reward of the selected action rises with $N$, while its clean reward stays near zero. (B) Safe task completion stays near 0\% with the imperfect verifier but reaches 100\% with a perfect verifier. Bands show 90\% Wilson confidence intervals over 10 episodes per point. Neither verifier violated any constraints.}
  \label{fig:rq1-frypan}
\end{figure}

Under over-responsive dynamics, human safety L2 safe completion falls from $82\%$ at $N=30$ to $70\%$ at $N=75$, while violations rise from $10\%$ to $18\%$ (Figure \ref{fig:rq1-suites}A--B).
Obstacle avoidance L0 tasks shows a larger reversal between $N=30$ and 50, followed by partial recovery at $N=75$ (Figure \ref{fig:rq1-suites}C--D). With output noise, soda handover task completion falls from $50\%$ at $N=5$ to $10\%$ at $N=75$, with no violations (Figure~\ref{fig:rq1-tasks}E).
For the banana placement task with under-responsive dynamics, violations rise from $40\%$ at $N=5$ to $80\%$ at $N=50$, although all episodes complete the task (Figure~\ref{fig:rq1-tasks}F). For the frypan placement task with output noise, increasing $N$ from 5 to 75 raises predicted reward from $0.1189$ to $0.1957$, while the clean reward of the same selected actions falls from $0.00356$ to $-0.00135$ (Figure~\ref{fig:rq1-frypan}).
Completion remains near zero despite the increase in proxy reward; it does not show an initial improvement followed by a decline.
Table~\ref{tab:rq1-changes} summarizes the selected execution changes.

\begin{table}[!htbp]
\centering
\small
\begin{tabular}{lrrrr}
\toprule
Setting & $N$ & Safe completion & Violations \\
\midrule
All 60 cases: imperfect & $1\to75$ & $43.8\%\to47.0\%$ & $10.3\%\to9.1\%$ \\
\midrule
Human L2: over-response & $30\to75$ & $82\%\to70\%$ & $10\%\to18\%$ \\
Obstacle L0: over-response & $30\to50$ & $34\%\to14\%$ & $4\%\to16\%$ \\
Soda: output noise & $5\to75$ & $50\%\to10\%$ & $0\%\to0\%$ \\
Banana: under-response & $5\to50$ & $60\%\to20\%$ & $40\%\to80\%$ \\
Frypan: output noise & $5\to75$ & $10\%\to0\%$ & $0\%\to0\%$ \\
\bottomrule
\end{tabular}
\caption{Performance impact under imperfect verification. First row: all 60 cases, pooled over the three verifier errors; with a perfect verifier, the same change in $N$ gives $43.8\%\to49.8\%$ safe completion and $10.3\%\to4.5\%$ violations. Other rows: groups and tasks chosen that show more reward hacking effects}
\label{tab:rq1-changes}
\end{table}

\subsection{RQ2: What leads to reward hacking in \texorpdfstring{best-of-$N$}{best-of-N} steering?}
\label{sec:rq2}

An unsafe action can be selected in two ways.
In an \emph{accepted unsafe choice}, the verifier predicts the violation but still selects the action because its task score outweighs the safety penalty.
In a \emph{missed unsafe choice}, the verifier predicts no violation, but the action violates a constraint when evaluated in the unperturbed simulator.
The first case comes from how safety and task progress are weighted and can occur even with a perfect verifier; the second comes from prediction errors and is the form of reward hacking we study.
To separate them, we compare each selected candidate's predicted violations with its violations in the unperturbed simulator.
We logged per-candidate scores for the 40 task-level cases at levels L0--L1 (ten repetitions each), so this analysis uses that subset rather than all 60 cases.
To isolate missed unsafe choices, we also consider \emph{safety-first selection}: it keeps only the candidates predicted safe ($\hat S_i=0$) and selects the one with the highest task score $\hat T_i$. If no candidate is predicted safe, it selects no action, and we exclude those states from the counts (Appendix \ref{app:mechanism}).

\begin{figure}[t]
  \centering
  \includegraphics[width=\linewidth]{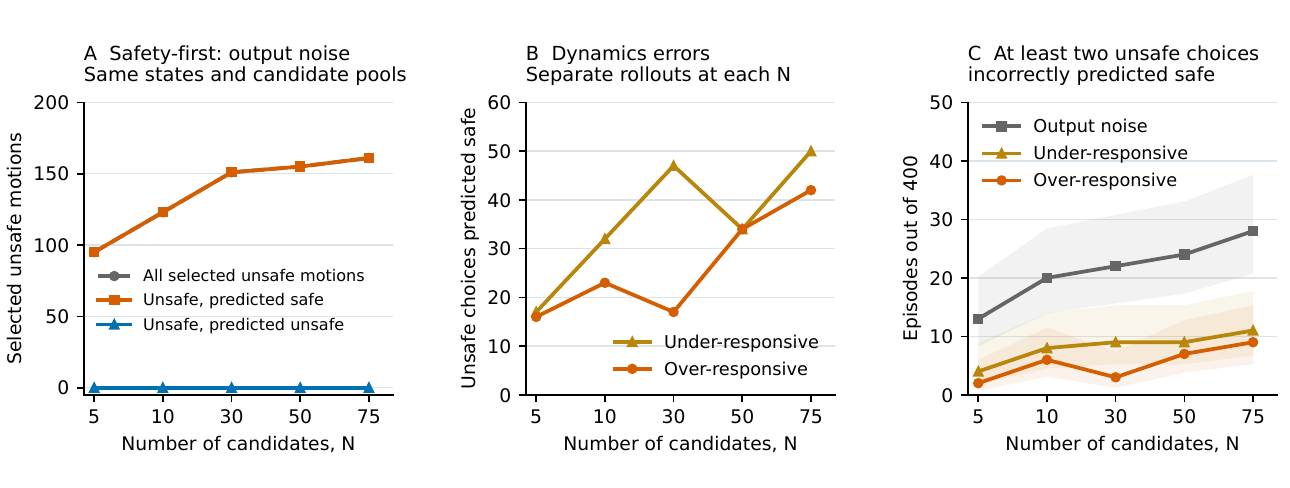}
  \caption{Increasing $N$ selects more unsafe actions that the verifier predicts to be safe. (A) Safety-first selection under output noise, at the same saved states and candidates: it never selects an action predicted to be unsafe, but selects more unsafe actions it misses as $N$ grows. (B) Missed unsafe choices under dynamics errors, from separate rollouts at each $N$. (C) Episodes (out of 400) with at least two missed unsafe choices. Bands show 90\% Wilson intervals.}
  \label{fig:mechanism}
\end{figure}

\paragraph{Missed unsafe choices increase with $N$.}
For output noise, we reuse the decisions from episodes run at $N=75$, where all 75 candidates were saved, and apply safety-first selection to only the first $N$ of them.
Because safety-first selection never picks an action predicted to be unsafe, every unsafe action it selects is a missed one.
The number of decisions with such an action rises from 95 at $N=5$ to 161 at $N=75$ (Figure~\ref{fig:mechanism}A): with more candidates, the verifier finds more actions that look safe and score well but in fact violate a constraint.
The same trend appears under dynamics errors, in the episodes run with weighted selection at each $N$: missed unsafe choices rise from 17 to 50 under under-response and from 16 to 42 under over-response (Figure~\ref{fig:mechanism}B).
Safety-first selection would not remove these either: a candidate predicted safe has $\hat S_i=0$, so if weighted selection picks it, it already has the highest task score among candidates predicted safe, and safety-first selection would pick the same one.
These choices also recur within episodes: the number of episodes with at least two missed unsafe choices rises with $N$ for all three error types (Figure~\ref{fig:mechanism}C).
Larger $N$ therefore increasingly selects unsafe actions that the verifier believes are safe, which no rule based on the verifier's own predictions can filter out.
Appendix~\ref{app:mechanism} gives the full procedure.

\subsection{RQ3: Can we mitigate reward hacking?}
\label{sec:rq3}

RQ2 showed that, as $N$ grows, selection increasingly favors candidates whose predicted scores are too optimistic.
This suggests two ways to reduce reward hacking: first, rely less on the single highest score, and second, change how the safety and task scores are combined. 
Both ideas have been used to mitigate inference-time reward hacking in LLMs \cite{khalaf2025inference,wang2026slop}, and we test each with $N=75$ fixed.

\subsubsection{\texorpdfstring{Soft best-of-$N$}{Soft best-of-N}}
\label{sec:soft}

Best-of-$N$ always takes the top-scoring candidate, and as $N$ grows, the top score is increasingly likely to belong to a candidate whose score is overestimated.
Soft selection hedges against this by spreading probability over several high-scoring candidates, so no single overestimated score decides the choice.
In LLMs, this kind of hedging reduces inference-time reward hacking~\citep{khalaf2025inference, wang2026slop}, and related work penalizes high scores that are uncertain~\citep{huang2025bestofn}.
Soft best-of-$N$ draws a candidate with probabilities based on the verifier scores:
\begin{equation}
  p_{\beta,c}(i)
  :=
  \frac{\exp\!\left(\beta\hat R_i/\sigma_c\right)}
       {\sum_{j=1}^{N}\exp\!\left(\beta\hat R_j/\sigma_c\right)},
  \label{eq:soft}
\end{equation}
where $\sigma_c$ is a fixed scale for verifier error $c$ that measures how much scores vary among candidates at the same decision (Appendix~\ref{app:soft}).
Smaller $\beta$ makes selection more uniform, and $\beta\to\infty$ recovers best-of-$N$.
We keep the original weights and evaluate $\beta\in\{0.5,2,8,16\}$ and best-of-$N$ on all 60 cases with ten repetitions.

\begin{figure}[t]
  \centering
  \includegraphics[width=\linewidth]{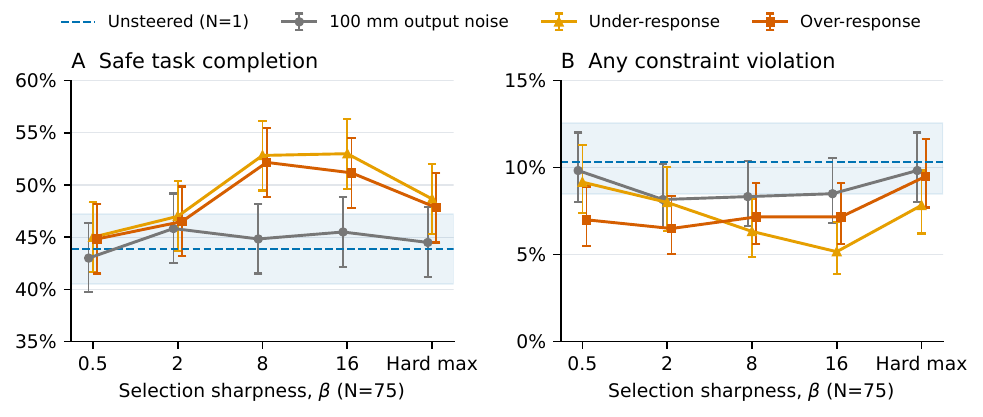}
  \caption{Soft best-of-$N$ at $N=75$ for different $\beta$; hard max corresponds to $\beta\to\infty$.
  (A) Safe task completion. (B) Episodes with any constraint violation.
  The dashed line and band show the unsteered policy ($N=1$), which has the same action distribution as uniform selection ($\beta=0$).
  Each point uses 600 episodes; intervals are pointwise 90\% Wilson intervals.}
  \label{fig:soft}
\end{figure}

Moderate $\beta$ improves observed outcomes over hard max under both dynamics errors (Figure~\ref{fig:soft}).
For under-responsive dynamics, $\beta=16$ raises safe completion from $48.7\%$ to $53.0\%$ and lowers violations from $7.8\%$ to $5.2\%$; under over-responsive dynamics, $\beta=8$ raises safe completion from $47.8\%$ to $52.2\%$ and lowers violations from $9.5\%$ to $7.2\%$.
These settings are about 8--9 points above the unsteered policy ($43.8\%$ safe completion, $10.3\%$ violations), while very soft selection ($\beta=0.5$) loses most of this gain.
This matches the intuition above: selection that is too sharp follows overestimated scores, while selection that is too soft ignores useful information in the scores.
Under output noise, the change is smaller: $\beta=2$ gives $45.8\%$ safe completion and $8.2\%$ violations, compared with $44.5\%$ and $9.8\%$ for hard max.
No single $\beta$ is best for both metrics under every error type, so soft selection offers a modest mitigation.

\subsubsection{Shared-weight calibration}
\label{sec:calibration}

Soft selection does not change how the scores are combined.
If verifier errors affect the safety and task scores differently, changing their weights can reduce the influence of the less reliable score.
Following SLOP~\citep{wang2026slop}, we fit one pair of weights $(w_S,w_T)$ for each verifier error, shared across tasks, to maximize the expected clean reward of the selected candidate on saved states from $N=75$ rollouts (Appendix~\ref{app:calibration}).
For dynamics errors, fitting uses three cases that are excluded from evaluation, leaving 57 held-out cases; for output noise, fitting uses all 60 evaluation cases.
The calibrated weights enter the softmax without the $\sigma_c$ scaling, so at $\beta=16$ their selection sharpness also differs from the original-weight setting in Section~\ref{sec:soft}.

\begin{figure}[t]
  \centering
  \includegraphics[width=\linewidth]{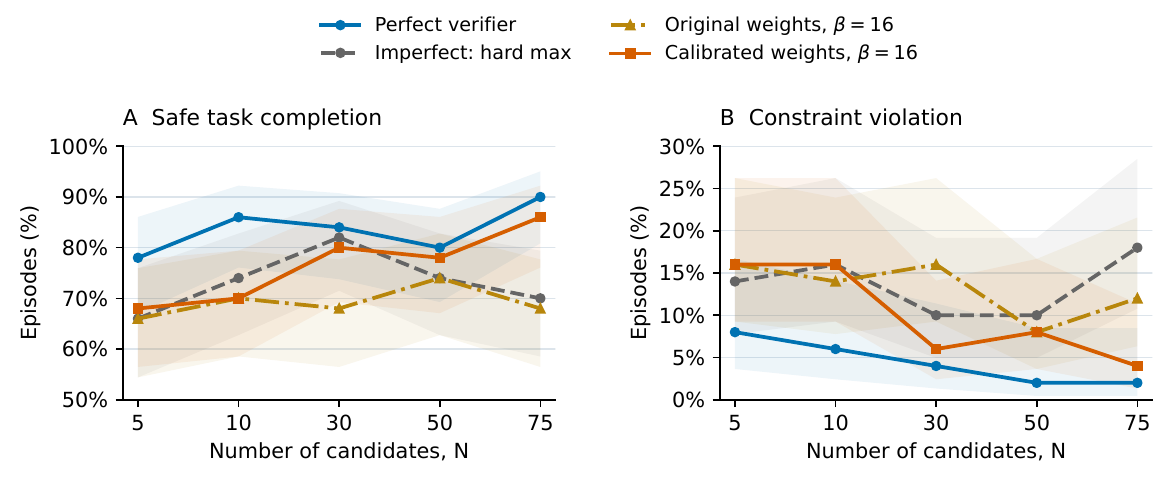}
  \caption{Calibration for human safety L2 tasks with over-responsive dynamics.
  All curves except the perfect verifier use the imperfect verifier, with the best-of-$N$ or soft best-of-$N$ ($\beta=16$) using the original or calibrated weights.
  (A) Safe task completion. (B) Constraint violations.
  Each point uses 50 episodes. Bands show pointwise 90\% Wilson intervals.}
  \label{fig:calibration-human-l2}
\end{figure}

For human safety L2 tasks with over-responsive dynamics, best-of-$N$ loses its gains at large $N$: from $N=30$ to $75$, safe completion falls from $82\%$ to $70\%$ and violations rise from $10\%$ to $18\%$ (Figure~\ref{fig:calibration-human-l2}).
With calibrated weights, safe completion instead rises from $80\%$ to $86\%$ and violations fall from $6\%$ to $4\%$.
At $N=75$, this gives $43/50$ safe completions and $2/50$ violating episodes, compared with $34/50$ and $6/50$ for soft best-of-$N$ with the original weights.

However, calibration is not a general fix.
Across the 57 held-out cases under over-responsive dynamics, safe completion barely changes ($52.3\%$ to $53.9\%$), and violations fall slightly ($5.4\%$ to $4.4\%$), while under-responsive dynamics cases become worse on both metrics.
In the under-output-noise case, even though the evaluation cases were used for fitting, calibration raises both safe completion and violations (Appendix~\ref{app:calibration}, Figure~\ref{fig:calibration}).

\section{Conclusion and Future Work}

In this paper, we show that best-of-$N$ test-time steering of VLA policies is vulnerable to inference-time reward hacking when the verifier is imperfect.
With a perfect verifier, increasing $N$ improves safe task completion, but with controlled errors in the verifier's simulation, larger $N$ widens the gap between predicted and clean reward and can erase these gains while increasing safety violations.
Our decision-level analysis explains why safety filtering alone does not prevent this: as $N$ grows, the selected unsafe actions are increasingly ones the verifier predicts to be safe.
Soft best-of-$N$ sampling and calibrated weights reduce this failure in some conditions, but neither is effective across all verifier errors.
Future work should test these effects on real robots, with other policies and learned verifiers, and under error types not studied here.

\clearpage
\bibliographystyle{plainnat}
\bibliography{references}

\begin{thebibliography}{20}
\providecommand{\natexlab}[1]{#1}
\providecommand{\url}[1]{\texttt{#1}}
\expandafter\ifx\csname urlstyle\endcsname\relax
  \providecommand{\doi}[1]{doi: #1}\else
  \providecommand{\doi}{doi: \begingroup \urlstyle{rm}\Url}\fi

\bibitem[Bakshi and Bose(2026)]{bakshi2026goodhart}
Soovadeep Bakshi and Apurba Bose.
\newblock Goodhart in the planner: Sampling-based control exploits rewards learned from human preferences.
\newblock \emph{arXiv preprint arXiv:2611.09670}, 2026.

\bibitem[Brohan et~al.(2023)]{brohan2023rt2}
Anthony Brohan et~al.
\newblock {RT-2}: Vision-language-action models transfer web knowledge to robotic control.
\newblock \emph{arXiv preprint arXiv:2307.15818}, 2023.

\bibitem[Cobbe et~al.(2021)Cobbe, Kosaraju, Bavarian, Chen, Jun, Kaiser, Plappert, Tworek, Hilton, Nakano, Hesse, and Schulman]{cobbe2021training}
Karl Cobbe, Vineet Kosaraju, Mohammad Bavarian, Mark Chen, Heewoo Jun, Lukasz Kaiser, Matthias Plappert, Jerry Tworek, Jacob Hilton, Reiichiro Nakano, Christopher Hesse, and John Schulman.
\newblock Training verifiers to solve math word problems.
\newblock \emph{arXiv preprint arXiv:2110.14168}, 2021.

\bibitem[Cui et~al.(2026)Cui, Zhang, Pang, Chi, Guo, Zhang, Xie, Jin, Mu, Yang, Yao, Zhan, Zhang, and Zhao]{cui2026liberosafety}
Rongxu Cui, Zongzheng Zhang, Jingrui Pang, Haohan Chi, Jinbang Guo, Saining Zhang, Shaoxuan Xie, Xin Jin, Yao Mu, Jiaolong Yang, Guocai Yao, Xianyuan Zhan, Ya-Qin Zhang, and Hao Zhao.
\newblock {LIBERO-Safety}: A comprehensive benchmark for physical and semantic safety in vision-language-action models.
\newblock \emph{arXiv preprint arXiv:2606.23686}, 2026.

\bibitem[Dai et~al.(2025)Dai, Liu, Bai, Liu, Wang, Su, Chen, Lin, and Wu]{dai2025rover}
Mingtong Dai, Lingbo Liu, Yongjie Bai, Yang Liu, Zhouxia Wang, Rui Su, Chunjie Chen, Liang Lin, and Xinyu Wu.
\newblock {RoVer}: Robot reward model as test-time verifier for vision-language-action model.
\newblock \emph{arXiv preprint arXiv:2510.10975}, 2025.
\newblock \doi{10.48550/arXiv.2510.10975}.
\newblock URL \url{https://arxiv.org/abs/2510.10975}.

\bibitem[Fei et~al.(2025)Fei, Wang, Shi, Dai, Cai, Qian, Ji, He, Zhang, Fei, Fu, Gong, and Qiu]{fei2025liberoplus}
Senyu Fei, Siyin Wang, Junhao Shi, Zihao Dai, Jikun Cai, Pengfang Qian, Li~Ji, Xinzhe He, Shiduo Zhang, Zhaoye Fei, Jinlan Fu, Jingjing Gong, and Xipeng Qiu.
\newblock {LIBERO-Plus}: In-depth robustness analysis of vision-language-action models.
\newblock \emph{arXiv preprint arXiv:2510.13626}, 2025.
\newblock \doi{10.48550/arXiv.2510.13626}.
\newblock URL \url{https://arxiv.org/abs/2510.13626}.

\bibitem[Gao et~al.(2023)Gao, Schulman, and Hilton]{gao2023overoptimization}
Leo Gao, John Schulman, and Jacob Hilton.
\newblock Scaling laws for reward model overoptimization.
\newblock In \emph{Proceedings of the 40th International Conference on Machine Learning}, volume 202 of \emph{Proceedings of Machine Learning Research}, pages 10835--10866, 2023.

\bibitem[Huang et~al.(2025)Huang, Block, Liu, Jiang, Krishnamurthy, and Foster]{huang2025bestofn}
Audrey Huang, Adam Block, Qinghua Liu, Nan Jiang, Akshay Krishnamurthy, and Dylan~J Foster.
\newblock Is best-of-n the best of them? coverage, scaling, and optimality in inference-time alignment.
\newblock In \emph{International Conference on Machine Learning}, pages 25075--25126. PMLR, 2025.

\bibitem[Khalaf et~al.(2025)Khalaf, Mayrink~Verdun, Oesterling, Lakkaraju, and du~Pin~Calmon]{khalaf2025inference}
Hadi Khalaf, Claudio Mayrink~Verdun, Alex Oesterling, Himabindu Lakkaraju, and Flavio du~Pin~Calmon.
\newblock Inference-time reward hacking in large language models.
\newblock In \emph{Advances in Neural Information Processing Systems}, volume~38, 2025.
\newblock \doi{10.52202/085713-2064}.

\bibitem[Kim et~al.(2024)Kim, Pertsch, Karamcheti, et~al.]{kim2024openvla}
Moo~Jin Kim, Karl Pertsch, Siddharth Karamcheti, et~al.
\newblock {OpenVLA}: An open-source vision-language-action model.
\newblock \emph{arXiv preprint arXiv:2406.09246}, 2024.

\bibitem[Kim et~al.(2025)Kim, Finn, and Liang]{kim2025fine}
Moo~Jin Kim, Chelsea Finn, and Percy Liang.
\newblock Fine-tuning vision-language-action models: Optimizing speed and success.
\newblock \emph{arXiv preprint arXiv:2502.19645}, 2025.

\bibitem[Kwok et~al.(2025)Kwok, Agia, Sinha, Foutter, Li, Stoica, Mirhoseini, and Pavone]{kwok2025robomonkey}
Jacky Kwok, Christopher Agia, Rohan Sinha, Matt Foutter, Shulu Li, Ion Stoica, Azalia Mirhoseini, and Marco Pavone.
\newblock {RoboMonkey}: Scaling test-time sampling and verification for vision-language-action models.
\newblock \emph{arXiv preprint arXiv:2506.17811}, 2025.

\bibitem[Lu et~al.(2025)Lu, Guo, Zhang, Zhou, Jiang, Gao, Tang, and Wang]{lu2025vla}
Guanxing Lu, Wenkai Guo, Chubin Zhang, Yuheng Zhou, Haonan Jiang, Zifeng Gao, Yansong Tang, and Ziwei Wang.
\newblock Vla-rl: Towards masterful and general robotic manipulation with scalable reinforcement learning.
\newblock \emph{arXiv preprint arXiv:2505.18719}, 2025.

\bibitem[Luo et~al.(2026)Luo, Xu, Wang, Zhou, and Zhang]{luo2026critichack}
Jiaxuan Luo, Xingguo Xu, Shanshan Wang, Yuhan Zhou, and Zhen Zhang.
\newblock Critichack: Evaluating visual rewards under robot policy optimization.
\newblock \emph{arXiv preprint arXiv:2610.02527}, 2026.

\bibitem[Nakamoto et~al.(2024)Nakamoto, Mees, Kumar, and Levine]{nakamoto2024vgps}
Mitsuhiko Nakamoto, Oier Mees, Aviral Kumar, and Sergey Levine.
\newblock Steering your generalists: Improving robotic foundation models via value guidance.
\newblock \emph{arXiv preprint arXiv:2410.13816}, 2024.
\newblock \doi{10.48550/arXiv.2410.13816}.
\newblock URL \url{https://arxiv.org/abs/2410.13816}.

\bibitem[{Physical Intelligence} et~al.(2025){Physical Intelligence}, Black, Brown, et~al.]{pi2025pi05}
{Physical Intelligence}, Kevin Black, Noah Brown, et~al.
\newblock $\pi_{0.5}$: A vision-language-action model with open-world generalization.
\newblock \emph{arXiv preprint arXiv:2504.16054}, 2025.

\bibitem[Stroebl et~al.(2026)Stroebl, Kapoor, and Narayanan]{stroebl2026limits}
Benedikt Stroebl, Sayash Kapoor, and Arvind Narayanan.
\newblock The limits of inference scaling through resampling.
\newblock In \emph{International Conference on Learning Representations (ICLR)}, 2026.

\bibitem[Wan et~al.(2026)Wan, Yue, Liu, Chu, Wang, Chen, Jiang, Zhu, Dong, and Zhu]{wan2026nofreechecker}
Yang Wan, Xihang Yue, Zhirui Liu, Ziyuan Chu, Shuxun Wang, Yuhan Chen, Xiaonan Jiang, Xukun Zhu, Yubo Dong, and Linchao Zhu.
\newblock No free checker: A survey of verifiers for robot policies.
\newblock \emph{arXiv preprint arXiv:2609.09250}, 2026.

\bibitem[Wang et~al.(2026)Wang, Liu, and Koike-Akino]{wang2026slop}
Ye~Wang, Jing Liu, and Toshiaki Koike-Akino.
\newblock Temper and tilt lead to {SLOP}: Reward hacking mitigation with inference-time alignment.
\newblock \emph{arXiv preprint arXiv:2605.13537}, 2026.

\bibitem[Zhang and Shah(2026)]{zhang2026veritas}
Mingtong Zhang and Dhruv Shah.
\newblock Visual verification enables inference-time steering and autonomous policy improvement.
\newblock \emph{arXiv preprint arXiv:2606.18247}, 2026.

\end{thebibliography}

\clearpage
\appendix

\section{Benchmark and suite comparisons}
\label{app:aggregate-results}
\begin{figure}[!htbp]
  \centering
  \includegraphics[width=\linewidth]{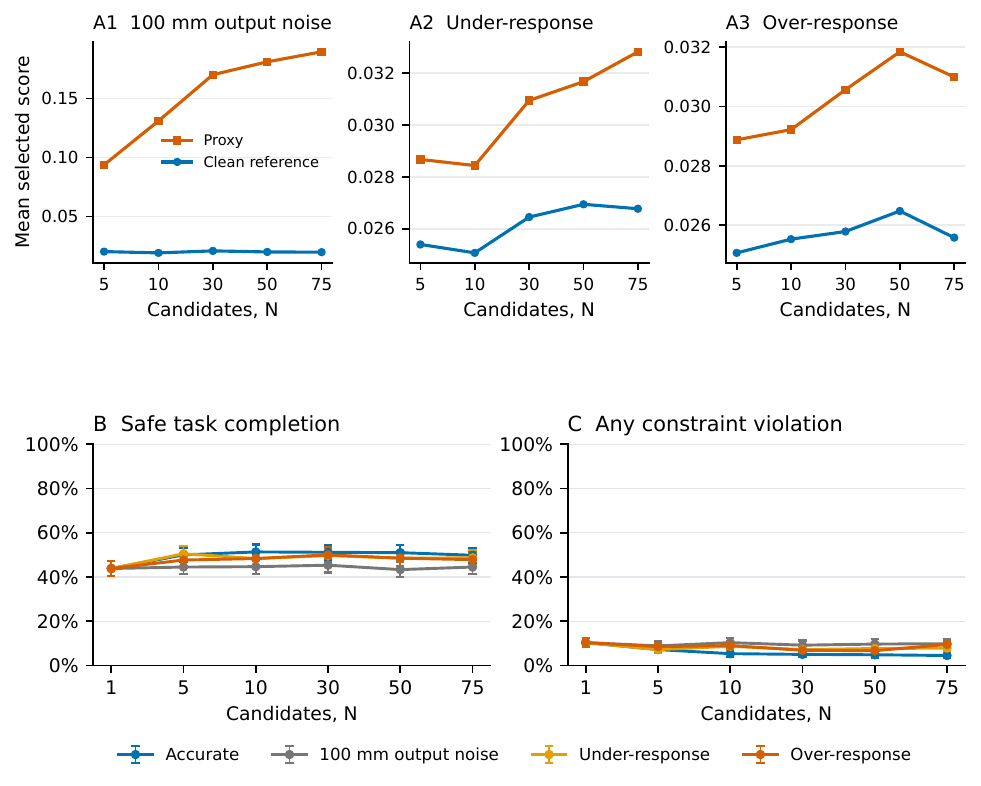}
  \caption{Benchmark-wide results under three verifier errors (60 cases, 600 episodes per point). (A1--A3) Predicted and clean reward of the same selected actions; note the different y-axis scales. (B) Safe task completion and (C) episodes with any constraint violation. Intervals are pointwise 90\% Wilson intervals.}
  \label{fig:main}
\end{figure}

This appendix gives benchmark-wide results and additional per-suite runs for Section~\ref{sec:rq1}.
We chose these groups after seeing the results, and they overlap, so they illustrate patterns rather than serve as independent confirmation.
With a perfect verifier, safe task completion rises from $43.8\%$ at $N=1$ to about $50\%$ for $N\geq5$, and violations fall from $10.3\%$ to $4.5\%$ at $N=75$ (Figure~\ref{fig:main}).
With an imperfect verifier at $N=75$, the mean predicted and clean rewards of the selected actions are $0.190$ and $0.020$ under output noise, $0.0328$ and $0.0268$ under under-response, and $0.0310$ and $0.0256$ under over-response.
The gaps are therefore $0.170$, $0.0061$, and $0.0054$; because scores vary much more under output noise, these gaps should not be compared directly across error types.
Safe task completion is $44.5\%$, $48.7\%$, and $47.8\%$, respectively, compared with $49.8\%$ for the perfect verifier, and violation rates are $9.8\%$, $7.8\%$, and $9.5\%$, compared with $4.5\%$.
More candidates thus raise the predicted reward without the gains in safe completion and safety seen with the perfect verifier.

\paragraph{Further reward evidence.}
For obstacle avoidance L0 with output noise, increasing $N$ from 30 to 75 raises the mean predicted reward of the selected action from $0.1823$ to $0.2023$, while its clean reward falls slightly, from $0.0229$ to $0.0217$.
Over the same range, safe completion falls from $32\%$ to $20\%$ (16 to 10 of 50 episodes), and violations rise from $14\%$ to $18\%$ (7 to 9).
In the two over-response groups in Figure~\ref{fig:rq1-suites}, by contrast, the mean predicted reward itself falls from $N=30$ to $75$, so their drop in outcomes is not by itself a case of predicted reward rising while clean reward falls.
These reward means are averaged over the states reached at each $N$, not over a fixed set of states.

\begin{figure}[!htbp]
  \centering
  \includegraphics[width=\linewidth]{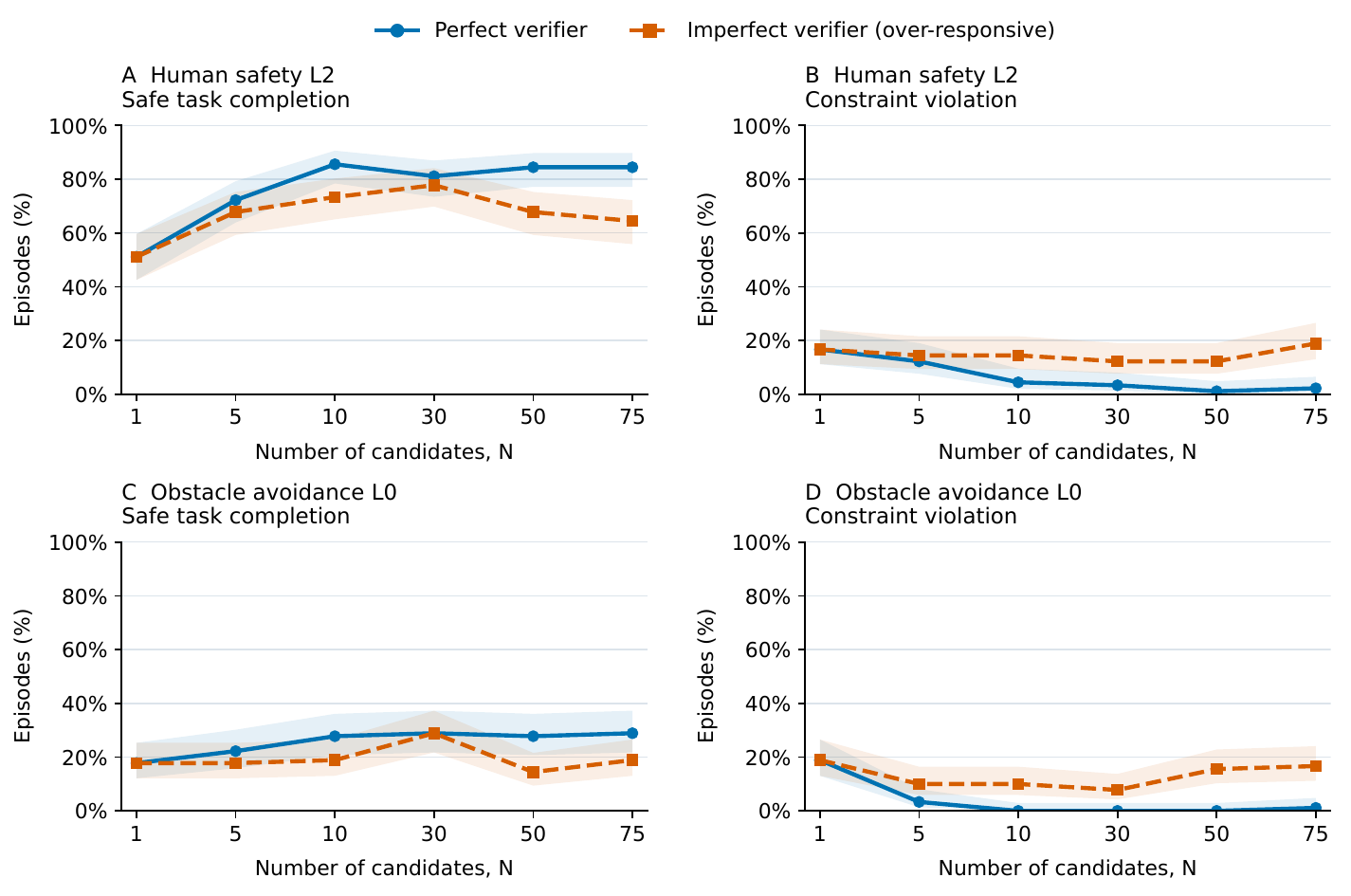}
  \caption{Over-responsive dynamics comparison from Figure~\ref{fig:rq1-suites}. (A,~B)~Human safety L2. (C,~D)~Obstacle avoidance L0.
Intervals are pointwise 90\% Wilson intervals.}
  \label{fig:rq1-suites-r18}
\end{figure}

\begin{figure}[!htbp]
  \centering
  \includegraphics[width=\linewidth]{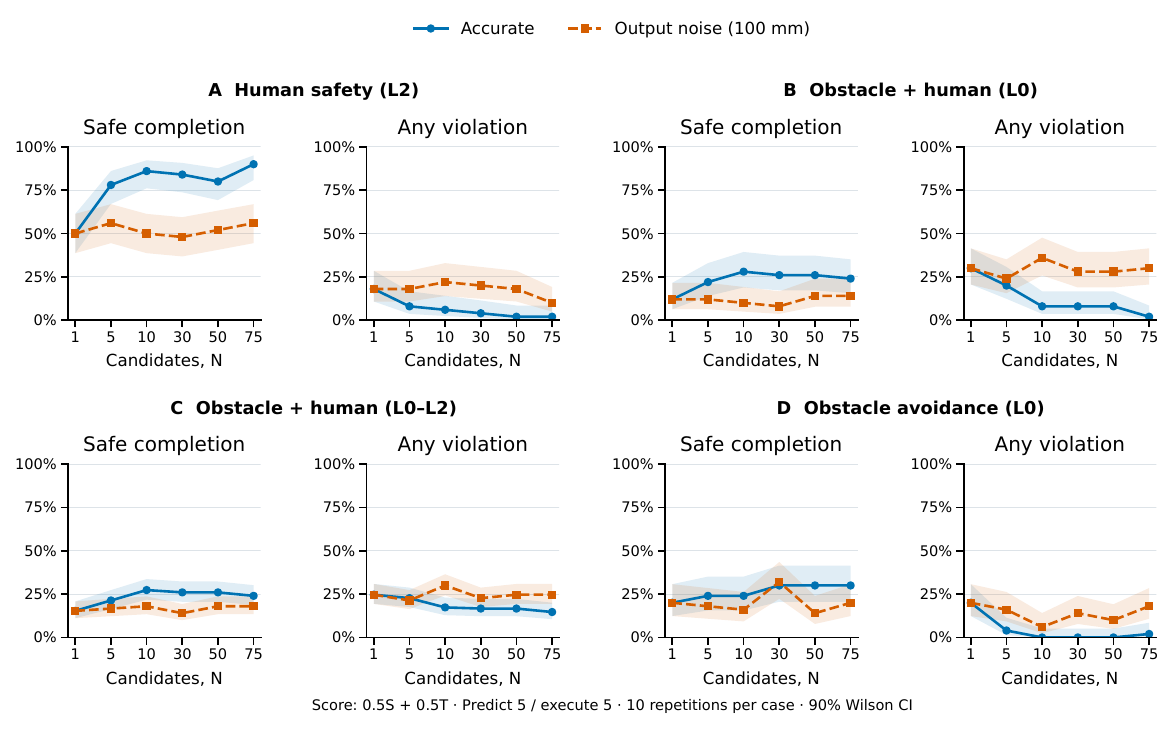}
  \caption{Safe task completion and constraint violations under output noise (100 mm noise in predicted object positions), compared with a perfect verifier.
  With the perfect verifier, completion rises and violations fall as $N$ grows; with output noise, both stay close to their $N=1$ values.
  Bands show pointwise 90\% Wilson intervals.}
  \label{fig:noise-suites}
\end{figure}

\begin{figure}[!htbp]
  \centering
  \includegraphics[width=\linewidth]{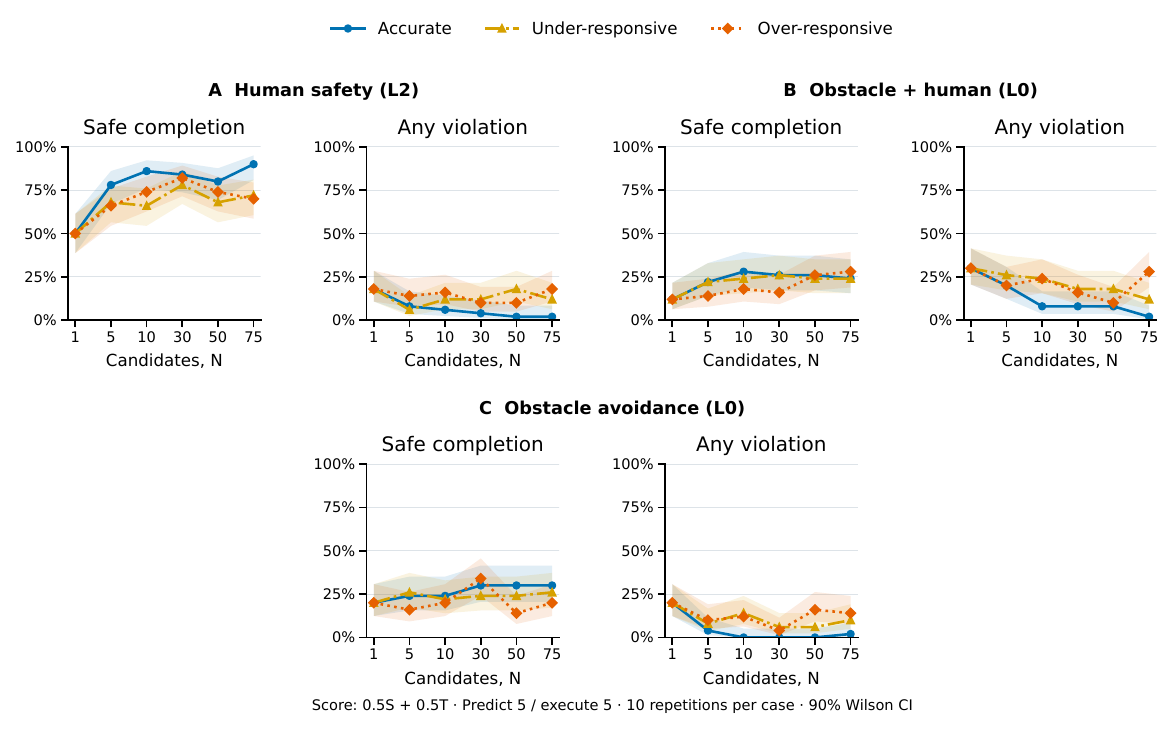}
  \caption{Safe task completion and constraint violations under under- and over-responsive dynamics, compared with a perfect verifier. The effect of dynamics errors varies across task groups and error types: in some groups, gains at moderate $N$ are lost at large $N$, while in others the curves stay close to the perfect verifier. Bands show pointwise 90\% Wilson intervals.}
  \label{fig:dynamics-suites}
\end{figure}

\paragraph{Additional runs.}
Figure~\ref{fig:rq1-suites-r18} repeats the over-response comparison with 18 repetitions per case, the largest number available for every case, so that each case counts equally.
The pattern for human safety L2 remains: with the imperfect verifier, safe completion falls from $77\%$ at $N=30$ to $64\%$ at $N=75$, and violations rise from $13\%$ to $20\%$, while the perfect verifier stays above $80\%$.
For obstacle avoidance L0 tasks, the pattern is weaker.
For the output-noise examples, we keep the original ten runs separate, because the additional runs fix the perturbation seed at zero instead of varying it across repetitions.
Not all curves decline steadily, either: with output noise, human safety L2 safe completion falls from $56\%$ at $N=5$ to $48\%$ at $N=30$ and returns to $56\%$ at $N=75$, and obstacle + human L0 rises from $8\%$ at $N=30$ to $14\%$ at $N=75$.
Figures~\ref{fig:noise-suites} and~\ref{fig:dynamics-suites} show the per-suite results for each error type.

\clearpage
\section{Decision-level safety analysis}
\label{app:mechanism}

We use saved records from the 40 cases at levels L0--L1, with ten repetitions per case and the original weights $w_S=w_T=0.5$.
A selected action is unsafe if it violates any constraint during its five control steps in the unperturbed simulator, and it is a missed unsafe choice if the imperfect verifier predicted no violations.

\paragraph{Output noise (Figure~\ref{fig:mechanism}A).}
For output noise, every candidate was also evaluated in the unperturbed simulator.
At each decision in episodes run with weighted selection at $N=75$, we apply safety-first selection to the first $N$ candidates, for $N\in\{5,10,30,50,75\}$.
Safety-first selection removes candidates with any predicted violation and then maximizes $\hat T_i$; ties go to the earliest candidate.
If no candidate is predicted safe, it selects no action; this happens at 268, 68, 3, 0, and 0 decisions for the five values of $N$.
We exclude these decisions from the counts, and we did not evaluate whether stopping there would be safe.
Because the decisions come from weighted selection, this analysis does not show which states a closed-loop safety-first controller would reach.

\paragraph{Dynamics errors (Figure~\ref{fig:mechanism}B).}
For dynamics errors, only the action chosen by weighted selection was evaluated in the unperturbed simulator, so we count missed unsafe choices among those actions in the episodes run at each $N$.
The number of decisions varies by less than 1\% across $N$, so the counts can be compared directly.
Safety-first selection would choose the same action at each of these decisions, but this does not show that it would be safe in closed loop.

\paragraph{Repeated missed choices (Figure~\ref{fig:mechanism}C).}
We count episodes with at least two decisions whose selected action is a missed unsafe choice; each condition and $N$ has 400 episodes.
This count need not equal the number of distinct contacts, and later choices may occur after an episode has already violated a constraint.

\clearpage
\section{Task-score details}
\label{app:task-score}

The task score follows the first unsatisfied goal predicate.
The manipulated object and target come from that predicate, with the benchmark's grasp and placement objects used as fallbacks.
Write $d_{eo}$ for the gripper-to-object distance, $d_{og}$ for the object-to-target distance, and $z_o$ for the object's height.
Superscripts $0$ and $h$ denote values before and after the candidate rollout.
The normalized approach, transport, and lift terms are
\begin{align}
 A_i&=\operatorname{clip}\!\left((d_{eo}^{0}-d_{eo}^{h})/0.08,-1,1\right),\\
 M_i&=\operatorname{clip}\!\left((d_{og}^{0}-d_{og}^{h})/0.10,-1,1\right),\\
 L_i&=\operatorname{clip}\!\left((z_o^{h}-z_o^{0})/0.05,0,1\right).
\end{align}
Distances are in meters; a distance term is zero when the required object or target is unavailable.
Let $b_g$ indicate a new grasp, $b_p$ a newly satisfied active goal predicate, and $I_h$ indicate that the object is grasped after the rollout and a target is available.
The dense score is
\begin{equation}
 D_i=\begin{cases}
  2M_i+L_i+3b_p, & \text{if the object was already grasped},\\
  A_i+2b_g+L_i+I_h M_i+3b_p, & \text{otherwise}.
 \end{cases}
\end{equation}
The normalized task score is then given by Equation~\ref{eq:task}.

\section{\texorpdfstring{Soft best-of-$N$ sampling settings}{Soft best-of-N sampling settings}}
\label{app:soft}

We estimate $\sigma_c$ once from saved $N=75$ candidate sets as the root-mean-square deviation of the proxy scores from their mean within each set.
We freeze $\sigma_c$ for all values of $N$ and $\beta$ and for closed-loop evaluation.
This normalization puts $\beta$ on a comparable scale across conditions with different reward magnitudes.
The scales for output noise, under-responsive, and over-responsive dynamics settings are $0.1102602$, $0.0245302$, and $0.0228821$, respectively.
Equation~\ref{eq:soft} changes only which of the $N$ candidates is selected.

\section{Calibration settings and fitted weights}
\label{app:calibration}

\subsection{Shared-weight calibration}

For a saved decision $p$ with 75 candidates, let $x_{pi}=(\hat S_{pi},\hat T_{pi})$ and let $g_{pi}=0.5S_{pi}^{\mathrm{clean}}+0.5T_{pi}^{\mathrm{clean}}$ be the clean reward.
Every candidate at a fitting decision must have a clean reward under dynamics errors; this requires replaying every candidate from the same initial simulator state with the original dynamics.
For weights $w=(w_S,w_T)$, fitting uses
\begin{equation}
 q_w(i\mid p)=\frac{\exp(w^\top x_{pi})}{\sum_j\exp(w^\top x_{pj})},\qquad
 J(w)=\frac{1}{|\mathcal P|}\sum_{p\in\mathcal P}\sum_i q_w(i\mid p)g_{pi}
       -\lambda\|w\|_2^2.
 \label{eq:fit}
\end{equation}
The expectation is over the candidates in each saved decision, not over new episodes.
We fit on decisions from $N=75$ rollouts and sample up to 30 decisions per task-level case to limit the influence of long episodes.

We compare unrestricted weights with nonnegative weights, and $\lambda\in\{0,10^{-5},10^{-4},10^{-3},10^{-2}\}$.
Each fit starts from $(1,1)$ and uses 200 Adam updates with learning rate $0.05$.
Configurations are compared by held-out expected clean reward, averaging within each task-level case before averaging across cases; lower unsafe-selection probability breaks ties.
We use five folds for output noise and three folds for dynamics, keeping each task-level case in one fold.
The selected configuration is then refitted on a new sample of up to 30 decisions per case.
All three conditions select nonnegative weights with $\lambda=0$.

The dynamics fitting cases are the soda-can handover in Human safety L0, the mug-and-microwave task in Obstacle avoidance L0, and the soup-and-cheese basket task in Obstacle avoidance + human L1.
These three cases are excluded from every method in the dynamics calibration comparison.
For output noise, fitting and evaluation cover the same 60 cases.

During closed-loop evaluation, we freeze the fitted weights and sample with probabilities proportional to $\exp(16w^\top x_i)$.
Because the weights are fit without this factor, selection at evaluation is sharper than the selection the weights were optimized for. The factor 16 is applied after fitting and is not part of Equation~\ref{eq:fit}.
The calibrated runs use score scale 1, whereas the original-weight $\beta=16$ runs use $\sigma_c$ in Equation~\ref{eq:soft}.
The results evaluate the complete procedure rather than isolating a change in the relative weights alone.

\begin{figure}[!htbp]
  \centering
  \includegraphics[width=\linewidth]{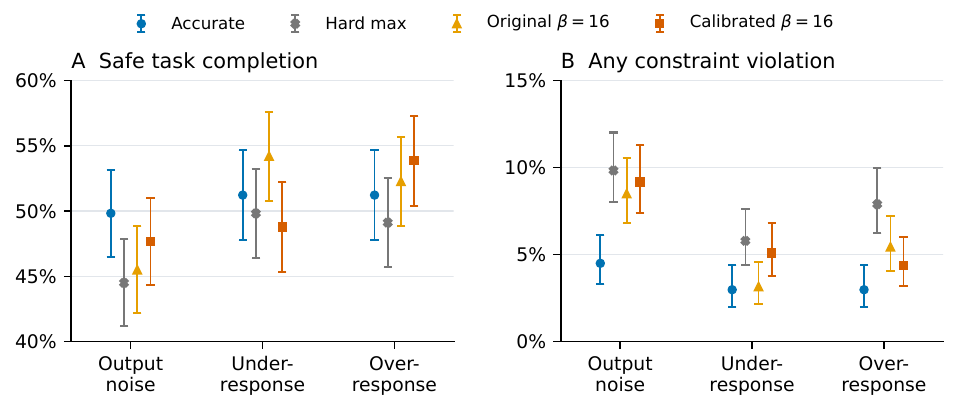}
  \caption{Aggregate calibration results at $N=75$. Output noise uses 600 episodes per method; dynamics use 570 after excluding the three fitting cases. All methods use ten repetitions per case, with 90\% Wilson confidence intervals.}
  \label{fig:calibration}
\end{figure}

For output noise, safe completion rises from $45.5\%$ to $47.7\%$, but violations also rise from $8.5\%$ to $9.2\%$.
For under-response, safe completion falls from $54.2\%$ to $48.8\%$ and violations rise from $3.2\%$ to $5.1\%$.
For over-response, safe completion rises from $52.3\%$ to $53.9\%$ and violations fall from $5.4\%$ to $4.4\%$.
Table~\ref{tab:weights} gives the fitted weights under output noise and the safety weight is zero.

\begin{table}[!htbp]
\centering
\small
\begin{tabular}{lrrrrr}
\toprule
Verifier error & Fitting cases & Available decisions & Used in final fit & $w_S$ & $w_T$\\
\midrule
Output noise & 60 & 20,035 & 1,800 & 0.0000 & 8.4305\\
Under-response & 3 & 315 & 90 & 6.8000 & 5.3504\\
Over-response & 3 & 304 & 90 & 5.7919 & 12.0803\\
\bottomrule
\end{tabular}
\caption{Shared weights fitted separately for each verifier error. Each decision has 75 candidates. Weights are neither constrained to sum to one nor normalized after fitting.}
\label{tab:weights}
\end{table}

\begin{table}[!htbp]
\centering
\small
\begin{tabular}{lrrr}
\toprule
Verifier error & Episodes & Safe completion, original $\to$ fitted & Violations, original $\to$ fitted\\
\midrule
Output noise & 600 & $273\to286$ & $51\to55$\\
Under-response & 570 & $309\to278$ & $18\to29$\\
Over-response & 570 & $298\to307$ & $31\to25$\\
\bottomrule
\end{tabular}
\caption{Episode counts behind the calibration comparison at $N=75$ and $\beta=16$. Both methods use ten repetitions per case.}
\label{tab:calibration-counts}
\end{table}

\end{document}